\documentclass{article}

\usepackage{arxiv}

\usepackage[utf8]{inputenc} 
\usepackage[T1]{fontenc}    
\usepackage{hyperref}       
\usepackage{url}            
\usepackage{booktabs}       
\usepackage{amsfonts}       
\usepackage{nicefrac}       
\usepackage{microtype}      
\usepackage{lipsum}

\usepackage{cite}
\usepackage[pdftex]{graphicx}

\graphicspath{{Figures/}}
\usepackage{xcolor}
\usepackage{amsmath}
\usepackage{tabularx,booktabs}
\usepackage{comment}
\usepackage[caption=false,font=footnotesize]{subfig}
\usepackage{multirow}
\newcommand{\angstrom}{\mbox{\normalfont\AA}}

\title{Protein Structured Reservoir Computing \\ 
for Spike-based Pattern Recognition}

\author{
  Karolos-Alexandros~Tsakalos \\
  Department of Electrical and Computer Engineering \\
  Democritus University of Thrace\\
  Xanthi, Greece \\
  \texttt{ktsakalo@ee.duth.gr} \\
   \And
  Georgios~Ch.~Sirakoulis \\
  Department of Electrical and Computer Engineering \\
  Democritus University of Thrace\\
  Xanthi, Greece \\
  \texttt{gsirak@ee.duth.gr} \\
   \And
  Andrew~Adamatzky \\
  Unconventional Computing Laboratory \\
  University of the West of England\\
  Bristol BS16 1QY, UK \\
  \texttt{andrew.adamatzky@uwe.ac.uk} \\
   \And
  Jim~Smith \\
  Computer Science Research Centre \\
  University of the West of England\\
  Xanthi, Greece \\
  \texttt{James.Smith@uwe.ac.uk} \\
}

\begin{document}
\maketitle

\begin{abstract}
Nowadays we witness a miniaturisation trend in the semiconductor industry backed up by groundbreaking discoveries and designs in nanoscale characterisation and fabrication. To facilitate the trend and produce ever smaller, faster and cheaper computing devices, the size of nanoelectronic devices is now reaching the scale of atoms or molecules --- a technical goal undoubtedly demanding for novel devices. Following the trend, we explore an unconventional route of implementing reservoir computing on a single protein molecule and introduce neuromorphic connectivity with a small-world networking property. We have chosen Izhikevich spiking neurons as elementary processors, corresponding to the atoms of verotoxin protein, and its molecule as a `hardware' architecture of the communication networks connecting the processors. We apply on a single readout layer, various training methods in a supervised fashion to investigate whether the molecular structured Reservoir Computing (RC) system is capable to deal with machine learning benchmarks. We start with the Remote Supervised Method, based on Spike-Timing-Dependent-Plasticity, and carry on with linear regression and scaled conjugate gradient back-propagation training methods. The RC network is evaluated as a proof-of-concept on the handwritten digit images from the \textcolor{black}{standard MNIST and the extended MNIST datasets and demonstrates acceptable 
classification accuracies in comparison with other similar approaches.}
\end{abstract}


\section{Introduction}
end of Moore's law indicates inability of CMOS technology to easily overcome the nanoscale dimensions \cite{Haron2008} due to quantum phenomena and, hence, the scaling of conventional transistors beyond gate lengths of 3~nm becomes almost unfeasible~\cite{meric2013,Tettamanzi2016}. Evolutionary prospects explored in computer science and information technology by imitating the functionality of mechanisms by which biological organisms process information might help us to deal with the scaling problem~\cite{Banzhaf1996,Li2017,Stepney2018}. The steadily growing disciplines that mainly contribute to the future of technology are Nanotechnology and Machine Learning. In particular, nanotechnology deals mostly with beyond Moore's law issues proposing tentative solutions, while the field of neuromorphic computing augments the classical computing principles by information processing in neural networks~\cite{Jackson2015, Tanaka2018, Pilarczyk2018, Lee2019}. 

Spiking Neural Networks (SNN) have been advanced in a sense of being currently comparable in neuromorphic behaviour to biological neural networks \cite{Paugam2012}. A relatively recent alternative approach called Reservoir Computing (RC) can be implemented on generic evolved or found Recurrent Neural Networks \cite{Verstraeten2009} and give rise to unconventional applications~\cite{nichele2017deep,konkoli2018reservoir,pilarczyk2018molecules} using nanoscale electronics such as nanoparticle materials~\cite{pilarczyk2018molecules} or even memristors~\cite{athanasiou2018using}. More specifically, RC systems are based on the principles of high-dimensional dynamical systems whose behaviour is interpreted as computation and are particularly suited for time varying and multidimensional signal classifications~\cite{Lukosevicius2012}. RC systems are designed in such a way that could transform a set of spike trains or sensory inputs into a spatiotemporal representation where structural feed-backs give rise to temporal memory states in the dynamics of the RC~\cite{Jaeger2004}. That collective RC state can be recognised by the readout layer of neurons which can learn to extract (in real time) its current state and past inputs. The key principle of RC systems is that its internal connectivity is kept fixed and the training process occurs only in the readout layer. This strategic design, by maintaining their internal circuitry, overcomes previous implementations of SNNs, due to the SNNs' lack of simple and efficient machine learning algorithms, and offers a practical training process. 

In order to utilise this kind of recursive systems, RC-based approaches traditionally are implemented by generating complex dynamical networks with typically randomised internal network topologies that are capable of creating various spatiotemporal states by which reservoirs store information from past inputs and produce responses which results from the reservoir’s ``memory" and present inputs~\cite{schrauwen2007overview}.

In this work, we consider the RC system as an Liquid State Machine (LSM) which is a type of reservoir computer that make use of spiking neural networks and proposed by Maass in 2002~\cite{Maass2002}. LSMs gained their attention due to their neuro-inspired architectures. Regular random structures of LSMs come with an accuracy trade-off. Plethora of studies suggests mechanisms with the ability to reconfigure the randomly created topology for achieving better accuracy~\cite{maass2011liquid}. This strategy is generally adopted in the literature, where specific LSM architectures are proposed for particular applications aiming to improve the corresponding accuracy. Various mechanisms have been continuously proposed by performing a network parameters optimization to achieve certain appealing features concerning the final network architecture. Such a similar approach is envisaged in the proposed work. Here, we propose an alternative approach of utilising bio-inspired networks and more specifically molecular assemblies so as to avoid the randomly generated networks that are made with a complex procedures, which sometimes take a long time to be executed and produce acceptable network properties. In this manner, this work attempts to use a real single molecular structure to create the internal connectivity of the RC system paving in a sense the way for alternative usage of real molecules for unconventional computing.

Molecules are nanomaterials fabricated by nature which have several desirable features as molecular networks describing their sub-nanometer scale, their hierarchical structure and small‐world character by their sparse star connectivity. They can also be considered as self-organising networks due to their property of molecular plasticity~\cite{Lamprecht2004}, as long as their structure is able to be reshaped and to be folded depending on the applied conditions in real-time~\cite{Lamprecht2004, Böde2007} as well as the ability to handle molecular dynamics through chemical and spectroscopy mechanisms makes molecules one of the most promising novel substrates for physical Reservoir computing~\cite{tanaka2019recent}. Their small-world character is based on their degree distribution which seems to follow the Poissonian distribution~\cite{atilgan2004small, bagler2005network}, i.e. macro-molecular assemblies have a much smaller number of hubs than the most of self-organised networks. The explanation for this deviation from the scale-free degree distribution lies in the limited restriction to simultaneous binding, while atoms may give up their bonds and bind to different local atoms during protein-folding and increasing the small-worldliness while the protein structure becomes progressively compact~\cite{Böde2007}. 

This allows us to introduce in this study two kinds of connectivity, the hard and the soft one which are determined by the macro-molecular bonding structure and the local 3D structure accordingly. In this way, we overcome the zero-clustering of the star connectivity consisting of the bonding structure and we provide an enhanced small world effect by scaling up the clustering mean coefficient due to local 3D interactions. Soft connectivity is described by a distance factor which allows the atom to interact with its local surroundings in 3D Euclidean space and it takes values in the scale of Angstrom (\angstrom). The major reason for engaging with soft connectivity is the fact that some attributes of small-world networks are shared on cortical networks~\cite{rubinov2010complex} which are significantly more clustered but have approximately the same characteristic path length in comparison with random networks~\cite{watts1998collective} which are widely used in Reservoir computing in order to produce its internal recurrent topology as mentioned before~\cite{tanaka2019recent,schrauwen2007overview}. 

\begin{figure}[!tbp]
  \centering
  \includegraphics[width=0.40\textwidth]{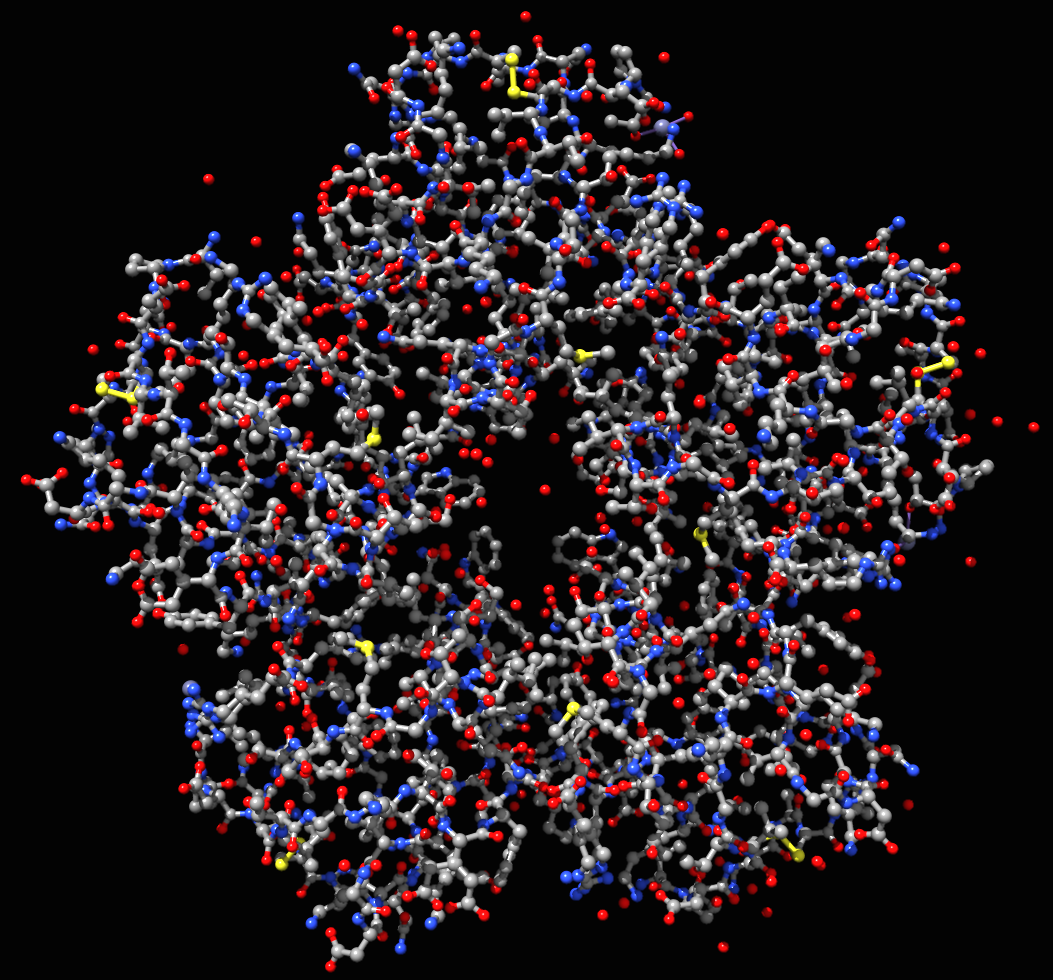}
  \caption{Verotoxin molecule under CPK colouring, colour convention for distinguishing individual atoms, determined in~\cite{Stein1992}. }
  \label{verotoxin}  \vspace{-15px}
\end{figure}

We demonstrate the structure of cell-binding B oligomer of Verotoxin1 molecule (VT-1) produced using X-ray diffraction intensities, with resolution 2.05\angstrom, obtained from E.coli~\cite{Stein1992} and shown in Fig.~\ref{verotoxin}. We used this molecular conformation as it’s been also demonstrated as an example with an excitable automata model~\cite{AdamatzkyVerotoxin} to preserve portability with our previous results in terms of Boolean gates realisation via interacting patterns of excitation, which can be illustrated as higher-dimensional transformations. 

We propose a novel approach by utilising already existing bio-inspired topologies, in particular molecular conformations, as opposed to randomly generated recurrent artificial topologies and we perform Reservoir Computing on a single molecule introducing the soft connectivity configuration that enables neuromorphic connectivity as it has a significant number of recurrent loops and presents a variety of spatio-temporal states. Non-linear dynamics and high-dimensionality is realised by introducing the Izhikevich neuromorphic model for the spiking neurons considered as elementary neuron-processors and corresponds to the atoms of VT-1 molecule, while their neuromorphic communication relies on the architecture based on both its hard molecular connectivity that is described by its chemical bonds and soft molecular connectivity described by its spatial local arrangement of atoms in euclidean space. We propose an unconventional reservoir computing approach by using this molecular conformation and utilizing its high-dimensional dynamics through the evaluation of a proposed STDP-based (Spike-timing-dependent-plasticity) pattern recognition training algorithm in a supervised fashion on a single output layer of spiking neurons, which is found mainly in echo state networks and not in liquid state machines we use here, in order to investigate its neuromorphic computing potential of dealing with well-known machine learning benchmarks such as solving the MNIST problem. 

We describe our approach during the RC setup as well as the calibration of the fine-tuned model parameters, which took place during the training phase, along with the evaluation phase of RC in terms of the MNIST problem, \textcolor{black}{where we addressed both the standard and the extended MNIST datasets}. Finally, we discuss our remarks and contributions as we compare it to other similar Reservoir Computing approaches and we give some suggestions for future works.

\section{Model}
\label{ProposedModel}
\subsection{Molecular Structure} \label{MolecularStructure}

\begin{figure}[thpb]
  \centering
  \includegraphics[width=0.45\textwidth] {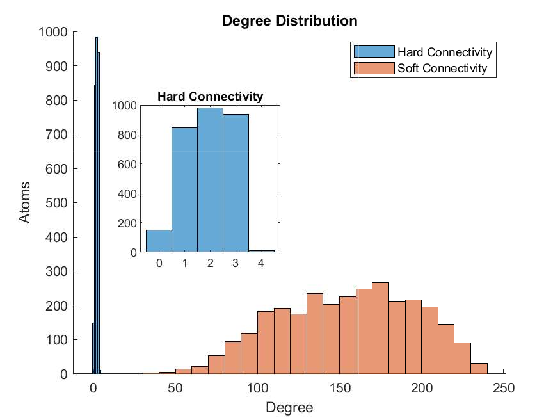}
  \caption{VT-1 molecular degree distribution for hard and soft connectivity under the distance factor of 10\angstrom. Hard neighbours of an atom represents the atoms which are connected to the atom through chemical bond while soft neighbours illustrates the atoms that are under a specific cut-off distance factor determined by the 3D structure of verotoxin~\cite{AdamatzkyVerotoxin}.}
  \label{verotoxinDegree}  
\end{figure}

\begin{figure}[thpb]
  \centering
  \includegraphics[width=0.45\textwidth]{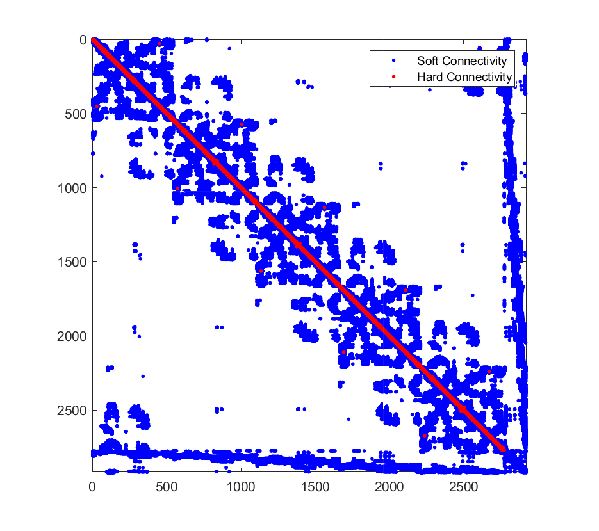}
  \caption{Sparsity pattern visualisation of the VT-1 Molecular Adjacency matrix for hard and soft connectivity accordingly. The strictly band-diagonal structure of the matrix is a result of the locality of internal connections.}
  \label{adjMATRIX}
\end{figure}

VT-1 is a pentamer protein of sixty-nine amino acids for each monomer and is thoroughly studied through its conversion to a non-directed graph whose vertices correspond to the atoms of the molecule and edges correspond to chemical bonds. In this study, the molecular hard connectivity represented by the graph's topology which has a set of 2,992 nodes and a set of 2,831 edges, respectively. Figure~\ref{verotoxin} illustrates the molecular hard connectivity of the VT-1 macro-molecule, and the Fig.~\ref{adjMATRIX} indicates the sparseness of the hard connectivity (red diagonal area) as long as the degree distribution is limited while the maximum degree observed is four for a few atoms as shown in Fig.~\ref{verotoxinDegree} (black distribution).   

\begin{figure}[thpb]
  \centering
  \includegraphics[width=0.45\textwidth]{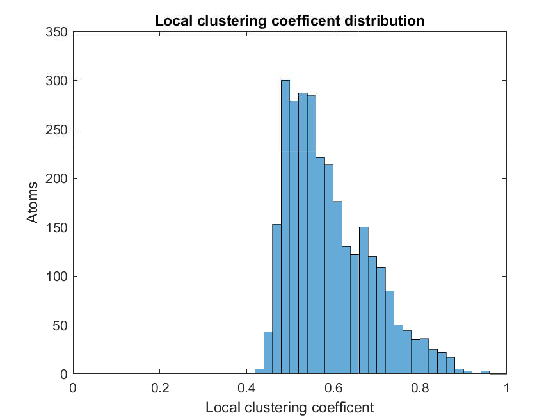}
  \caption{Local clustering coefficient distribution of the VT-1 molecule when we create soft connectivity with distance factor 10\angstrom.}
  \label{localclustering}  
\end{figure}

Various studies regarding RC systems have showed that Orthogonal Recurrent connectivities achieve better results for larger networks, similar to the proposed architecture, in comparison with other regular random connectivities~\cite{mayer2017orthogonal,boedecker2012information}. In order to utilize this feature we introduce the soft connectivity of the molecular structure.  Let $\delta$ be an average distance between two hard-neighbours, for F-actin $\delta=1.43$ units. Let $w(s)$ be nodes of actin molecule that are at distance not exceed $\rho$, in Euclidean space, from node $s$. We call them soft neighbours because their neighbourhood is determined by 3D structure of the molecule. This configuration gives us some attractive features like the increased fractal-like connectivity density along with orthogonality as shown in Fig.~\ref{adjMATRIX} (black area) as well as the orange right-shifted degree distribution that becomes increasingly evident as Poissonian-like distribution as the Fig.~\ref{verotoxinDegree} indicates.

Looking carefully at the adjacency matrix, five orthogonal clusters are created within the topology, where in each cluster there is increased connectivity and there are also connections between the nearby clusters. Beyond this observation, is should be noticed that there is a region, shown at the bottom of Fig.~\ref{adjMATRIX}, which is connected across the network, representing the role of hubs and enabling the small-world effect~\cite{watts1998collective}. This feature has been claimed as biological feasible~\cite{achard2006resilient,bassett2006small,varshney2011structural} through the intuition that a small-world network integrates information from many regions. These types of networks are resilient to inherent faults\/damages of the reservoir connectivity~\cite{achard2006resilient,hazan2012topological}, which conveys a clear message that future implementations of molecular devices could be robust to defects for various neuromorphic applications.

Along with that, the most intriguing properties of soft connectivity are associated with the clustering coefficient and the average path length of the network. In the case of hard connectivity, it is well-known that proteins are a single chain of amino acids which corresponds to star connectivity due to their bonding structure and have zero-clustering. In contrast, when we include soft connectivity, we notice that we have high clustering as shown in Fig.~\ref{localclustering}, which represents the local clustering coefficient distribution, and in particular, the average clustering coefficient of the network is $C=0.60$ while the average path length becomes $L=4.93$ when the distance factor is 10\angstrom. As the distance factor is increased to 20\angstrom, we notice that the average clustering coefficient increases slightly to $C=0.69$ and the average path length decrease to $L=1.88$. In this way, we utilise the VT-1 molecular structure integrating on each atom a neuromorphic behaviour. Compared with the SNN approach, which is a general paradigm used in modelling biological neural networks for machine learning purposes, RC are a not-so-general-paradigm; in the latter, the goal is to understand how ensemble neurons process information. Nevertheless, in both cases, many different neuron models can be considered, in which the information is encoded in temporal distance between the consecutive action potentials. 

\subsection{Neuron Model}

In literature, various neuromorphic behaviours are modelled at different levels of abstraction, ranging from the most biologically realistic Hodgkin-Huxley (HH) model to the simplest and most computationally efficient Leaky Integrate-and-Fire (LIF) model. In our case, we use the Izhikevich neuron model \cite{Izhikevich2004, Izhikevich2010} since it offers a good compromise between a biological accuracy and the computational efficiency, as well as it manages to produce several kinds of spike and burst patterns observed in biological neurons by the proper choice 
of only four variables~\cite{ZHANG2018}. 

The Izhikevich model is described by the following two-dimensional system of differential equations

\begin{eqnarray}
&\dot{v}(t) &= 0.04v^2(t)+5v(t)+140-u(t)+I(t) \\
&\dot{u}(t) &= a(bv(t)-u(t)) \\
&I_i(t) &=\sum^{fired}_{j} I_os_{ij} + I_i^{in}(t)  
\end{eqnarray}
with after spike resetting 
\begin{eqnarray}
if \,\, v(t) \geq 30mV \,\,
   \begin{cases} 
      v(t) \rightarrow c \\
      u(t) \rightarrow u(t)+d 
   \end{cases}  
\end{eqnarray}

\noindent where \emph{v} is the membrane potential, \emph{u} is a recovery variable that contains the dynamics of ions-channels, \emph{I} represents current stimulation of the neuron and \emph{a}, \emph{b}, \emph{c} and \emph{d} are dimensionless parameters. Regarding the reset phase, if the membrane potential reaches the firing threshold, then the neuron generates a spike which affects only the adjacent neurons. The variable \emph{t} refers to the simulation time that is distinguished in 1\emph{ms}. The variable $s_{i,j}$ denotes the synaptic weight, between the post-synaptic neuron \emph{i} and the pre-synaptic neuron \emph{j}, that varies within the range of (0,1) and creates a specific spike which is multiplied by the $I_o$ current amplitude to give the appropriate interconnection current value of the order of \emph{pA}. This equation represents the sum up of the synapses of the presynaptic neurons that fired together with the addition of an external excitation current $I_i^{in}(t)$ to neuron \emph{i}. 

\subsection{Network Architecture}
\subsubsection{Input Encoding}

\begin{figure}[!htpb]
    \centering
    \includegraphics[width=.6\linewidth]{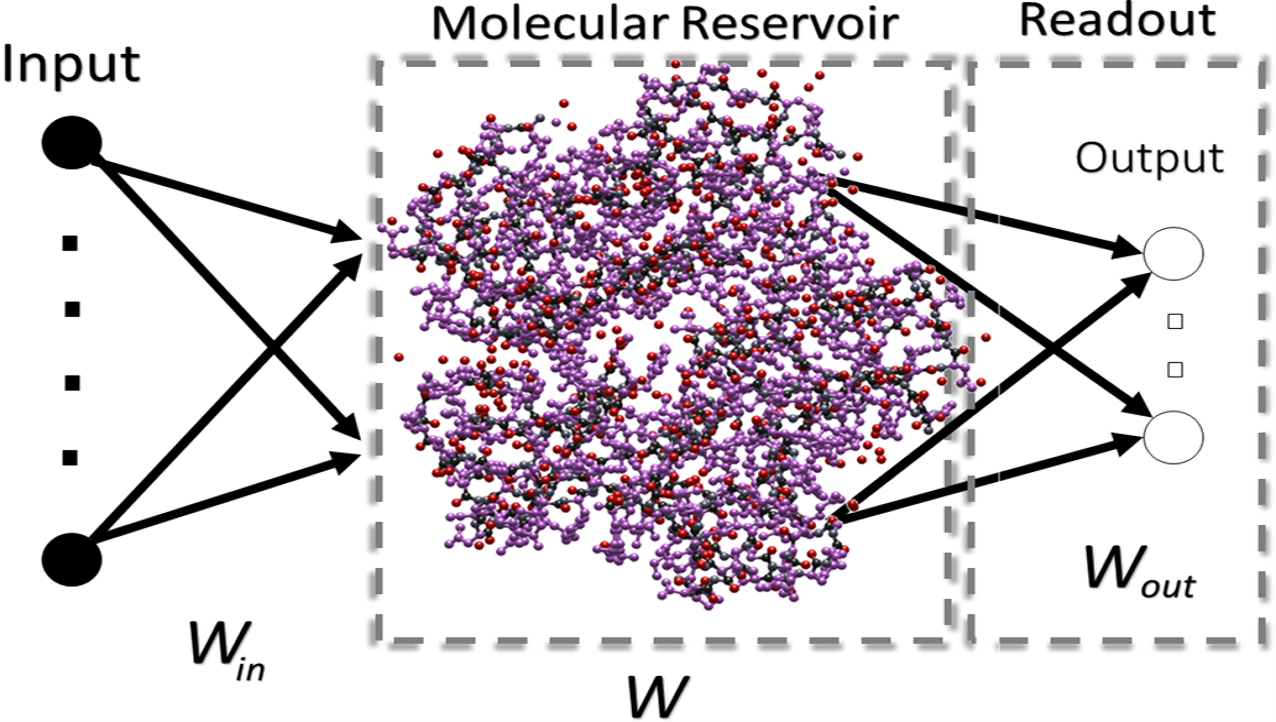}
    \caption{\textcolor{black}{Protein Structured Reservoir Computing Model.} }
        \label{reservoir}   
\end{figure}

\begin{figure*}[thpb]
\subfloat[\label{neuromorphic_response_a}]
{\includegraphics[width=0.98\textwidth]{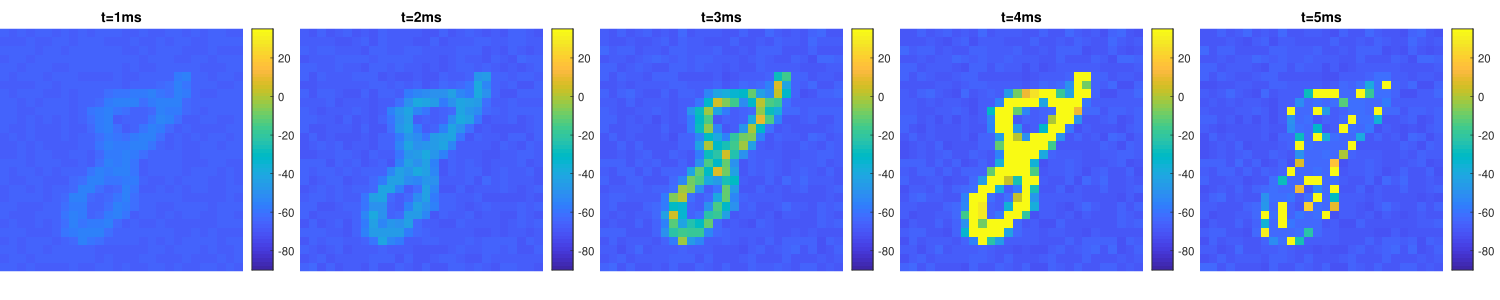}}

\subfloat[\label{neuromorphic_response_b}]
{\includegraphics[width=0.98\textwidth]{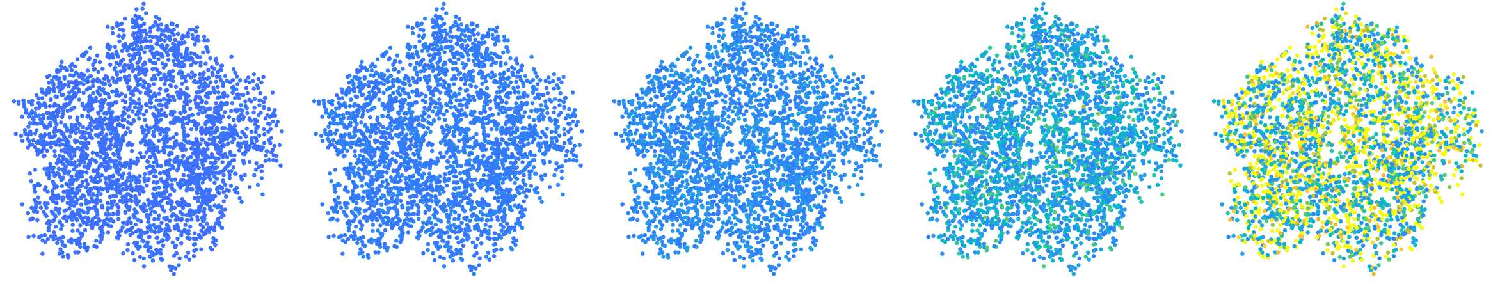}}
\caption{Spatio-temporal representation of the neuromorphic activity in terms of membrane potential (mV) (a)~from the 28$\times$28 Input Layer grid through external stimulation of a random 28$\times$28 image from the standard MNIST dataset down-scaled on the range of $(0,1)$ and 
(b)~from the VT-1 molecular-based RC response over 5 time-steps, 1~ms each, from the Input Layer action potentials.}
\label{neuromorphic_response}  
\end{figure*}

The RC approach can be divided into three parts, the input interface $W_{in}$, the RC network $W$ and the readout interface $W_{out}$ as depicted in Fig.~\ref{reservoir}. The input interface concerns the transformation of the input stimulation to time-varying input spike trains 
that excite the RC system. At an early stage, the RC was directly stimulated with the properly-scaled input pixels' intensity in the form of current inputs. Initially, $784$ ($28\times28$) random RC nodes were directly stimulated. However, due to limited accuracy, various combinations of direct input stimulation to RC system were performed. Starting with the all around combination in which the image is stimulated sequentially about four times on the network. Even so, spatio-temporal transformation of the input does not take place, i.e. the output was highly correlated with the input, which resulted in better but still no comparable performances. 
\textcolor{black}{As a result, this input spike generation is realised through transforming the input image pixels' to spike trains by the input layer of $28\times28$ uncoupled Izhikevich-based neurons. The images' pixel values are pre-processed and converted from $[0,255]$ range to $[0,1]$ range of input currents. This is considered as the stimulation of the input layer which are turned to Izhikevich-based spike-trains, with proportional firing rates to the pixel's intensity. After that, each RC neuron receives a single randomly selected spike-train from a single neuron of the input layer.} Figure~\ref{neuromorphic_response_a} illustrates the neuromorphic activity of the 28$\times$28 input layer in terms of the membrane potential reflecting on a input image for an interval of five time-steps of $1$~ms each and Fig.~\ref{neuromorphic_response_b} shows the response of the molecular-based RC network in terms of the membrane accordingly. We notice that action potentials of the input layer affect significantly the neuromorphic activity on the RC layer at the next time-step. 

\subsubsection{Reservoir}
RC system consists of a 3D structured locally connected network of spiking neurons and is usually randomly created using biologically inspired parameters. In this study, we use the molecular connectivity as the RC network and the responses from all neurons are projected to the next output layer, where the actual training is performed through supervised learning algorithms to recognize instantaneous spike-based spatio-temporal patterns within the RC. 

In the view of avoiding chaotic dynamics, we introduce two kinds of neurons with different type of spiking behaviour, the excitatory and inhibitory neurons~\cite{Izhikevich2007} with a 4:1 ratio of excitatory to inhibitory neurons as in the mammalian cortex~\cite{Druga2009}. To achieve this, we employ a regular behaviour with the parameters shown in Tab.~\ref{table:ExcInh} for excitatory and inhibitory neurons. From the VT-1 molecule we select oxygen and hydrogen atoms as inhibitory neurons which are about $26\%$ among all atoms and they are distributed throughout the network, while the rest atoms are selected as excitatory neurons. 

\begin{table*}[!thpb]
    \centering
    \caption{Parameters of the neuromorphic model of Izhikevich used for the molecular-based RC system behaviour for each neuron}
    \begin{tabular}{m{2em} c c c m{2em}}
     \toprule \toprule\\
     & Izhikevich       & Excitatory     & Inhibitory&\\ 
     & Parameters       & Neurons        & Neurons&\\ [1ex] 
     \midrule\\ 
     & $a$ & 0.02         & 0.02 to 0.10   & \\
     & $b$ & 0.20         & 0.20 to 0.25   & \\
     & $c$ & -65 to -50   & -65            & \\
     & $d$ & 2 to 8       & 2              & \\[1ex]
     \bottomrule \bottomrule
    \end{tabular} 
    \label{table:ExcInh}  
\end{table*}

\subsubsection{Learning Mechanism}

We adopted spike-timing-dependent plasticity (STDP) which indicates  \textcolor{black}{the correlation of the firing of the pre-synaptic neuron $n_{k}^{i n}(i)$ and post- synaptic neuron $n_{j}^{o}(i)$. This plasticity mechanism suggests that since the pre-synaptic neuron always fires just before the post-synaptic neuron the firings are correlated and the synaptic weight between them should be increased; otherwise, if a pre-synaptic neuron fires just after the post-synaptic neuron the synaptic weight should be decreased.} The weight adjustment depends on the firing times of the pre- and post-synaptic neurons accordingly~\cite{Dan2004}. One of the most widely used STDP-based training algorithms, which adapts synaptic weights according to the mechanisms of \emph{STDP} and \emph{anti-STDP} is the \emph{Remote Supervision Method (ReSuMe)} \cite{Ponulak2006,Ponulak2005}. 

Two key features of ReSuMe learning are employed: the \textbf{Remote Supervision} and the \textbf{Learning Window}. The remote supervision concerns a remote ``teacher" neuron for each output neuron, or more precisely the use of remote firing times. Regarding this, the synaptic weights depend, not only on the correlation between the firing times of neurons $n_{k}^{i n}(i)$ and $n_{j}^{o}(i)$, but also on the correlation of firing times of neuron $n_{k}^{i n}(i)$ and remote neuron $n_{j}^{d}(i)$ 
The function called the learning window, determines the correlation between the firings and thus the synaptic weight adjustment. 

The ReSuMe learning is executed by the following cost function derivative:

\begin{equation}
\frac{\mathrm{dw_{k i}(t)}}{\mathrm{d} t} =\eta\left[S^{d}(t)-S^{o}(t)\right]\left[1+\int_{0}^{\infty} W(s) S^{i n}(t-s) \mathrm{d} s\right]
\end{equation}

\noindent where $S^{d}(t), S^{i n}(t)$ and $S^{o}(t)$ are the precise desired pre- and post-synaptic spike trains, respectively, the constant $\eta$ represents the non-Hebbian contribution to the weight changes and the learning rate, while function $W(s)$ of a time delay $s=t-t^{fired}_{in}$ between the correlated spikes is known as a learning window. The shapes of $W(s)$ applied in ReSuMe are similar to the ones used in STDP models and can be represented by the following equation. In this way, over time ReSuMe Learning aims to get the desired and the output spike trains even closer to each other, i.e. to have a simultaneous firings.

\begin{equation}
W(s)=\left\{\begin{array}{ll}{+A_{+} \cdot e^ {-s / \tau_{+}}} & {\text { if } \quad s \geq 0} \\ {-A_{-} \cdot e^ {s / \tau_{-}}} & {\text { if } \quad s<0}\end{array}\right.
\end{equation}
\noindent with amplitudes $A_{+}, A_{-} \geq 0$ and time constants $\tau_{+}, \tau_{-}>0$ of the positive and negative parts of the learning window, respectively~\cite{ponulak2008analysis}.  

\subsubsection{Training and Optimization}

In our case, we aim to reduce the training complexity and in particular to cause the firing of neurons in the shortest possible time interval. Thus, we could provide a rich neuromorphic activity to proceed to readout layer training accordingly in a short period of time.

\begin{figure*}[!htb]
    \includegraphics[width=1\textwidth]{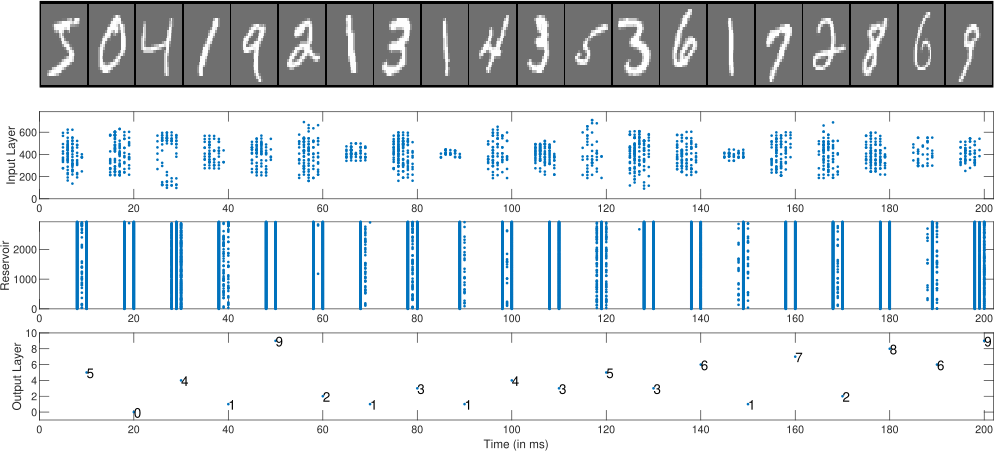}
    \caption{Spatio-temporal spike-based pattern representation of the proposed Molecular-based LSM. We present the raster diagrams regarding the spike-trains obtained by the input layer, the spike-trains of the Molecular-based LSM, as well as the time-dependent spikes of the output layer. Resume learning is performed under \textcolor{black}{a learning duration $T\text{=}10~\text{ms}$} for each digit and the classifier neurons are trained to spike right at the last ms of the learning duration.}
    \label{NetworkResponse} 
\end{figure*}

The RC system except the recurrent structure also consists of a single output classifier layer with spiking neurons which have also been integrated by the neuromorphic model of Izhikevich. Since the RC is driven by the handwritten digits from the MNIST-based datasets, ten spiking readout neurons (labelled 0-9) represents the classified digit value of the input image, schematically illustrated in Fig.~\ref{NetworkResponse}. This dataset consists of thousands of handwritten images which have $28\times{28}$ pixels in grayscale.

\textcolor{black}{Fig.~\ref{NetworkResponse} presents the RC framework, in which each input image is converted to spike-trains through the input layer and presented to RC network for a specific time interval, the so-called learning duration $T$.} The raster diagram represents the exact spike timings of the input layer, the RC network and the output layer response for each image after the training phase \textcolor{black}{with a learning duration of $T\text{=}10~\text{ms}$. In terms of improving the network overall accuracy and preventing over-fitting on the training dataset, smaller values of \emph{T} are more efficient, which often results in fewer iterations to learn the RC dynamics produced from each image. We reduce the training complexity and cause the firing of neurons in the shortest possible time interval to provide a rich neuromorphic activity on the output readout layer and train it in a short period of simulation time. The network overall accuracy is optimised with ReSuMe learning under a learning duration of $T\text{=}6\text{ms}$.}

In order to evaluate the spike-based responses obtained from the RC network, aiming to present the perspectives of our approach, well-known classifiers are employed to utilize the molecular-based encoded information. Multiple linear regression algorithm is used as classifier on the readout layer. By utilizing the spike-based information encoded through the RC system, we encode the raster diagram as a RC state vector $X$, which represents the overall spiking activity of the network and 
is applied to classifier. Regarding the SCG training the classifier consists of a multi-layer perceptron, while multiple linear regression classifier includes a layer of ten neurons that use the sigmoid activation function described by the following equations:

\begin{equation}
\alpha_{\theta}(\mathbf{x})=f\left(\boldsymbol{\theta}^{\mathrm{T}} \cdot \mathbf{x}\right)
\end{equation}\begin{equation} f(\mathbf{x})=\frac{1}{1+\mathrm{e}^{-\mathbf{x}}}\end{equation} 
where $f(x)$ is the sigmoid function and $\boldsymbol{\theta}^{\mathrm{T}}$ is the readout weights that are trained through linear regression. In this manner, the cost function is described as 
\begin{equation}J(\boldsymbol{\theta})=\frac{1}{2n} \sum_{i=1}^{n}\left(\mathbf{y}_{d} -\alpha_{\theta}\left(\mathbf{x}^{(i)}\right)\right)^2\end{equation} 
where $n$ is the number of images, $y_d$ is the desired output and the $\alpha_{\theta}\left(\mathbf{x}^{(i)}\right)$ is the observed output. 

The readout layer is trained using gradient descent, as described below, in order to minimise the cost function. 
\begin{equation}
\frac{\partial J(\boldsymbol{\theta})}{\partial \boldsymbol{\theta}_{j}}= \frac{1}{n} \sum_{i=1}^{n}\left(\mathbf{y}_{d}-\alpha_{\theta}\left(\mathbf{x}^{(i)}\right)\right) \mathbf{x}^{(i)}
\end{equation}

\begin{figure*}[!htb]
   \subfloat[\label{stdp}]{%
      \includegraphics[width=0.33\textwidth]{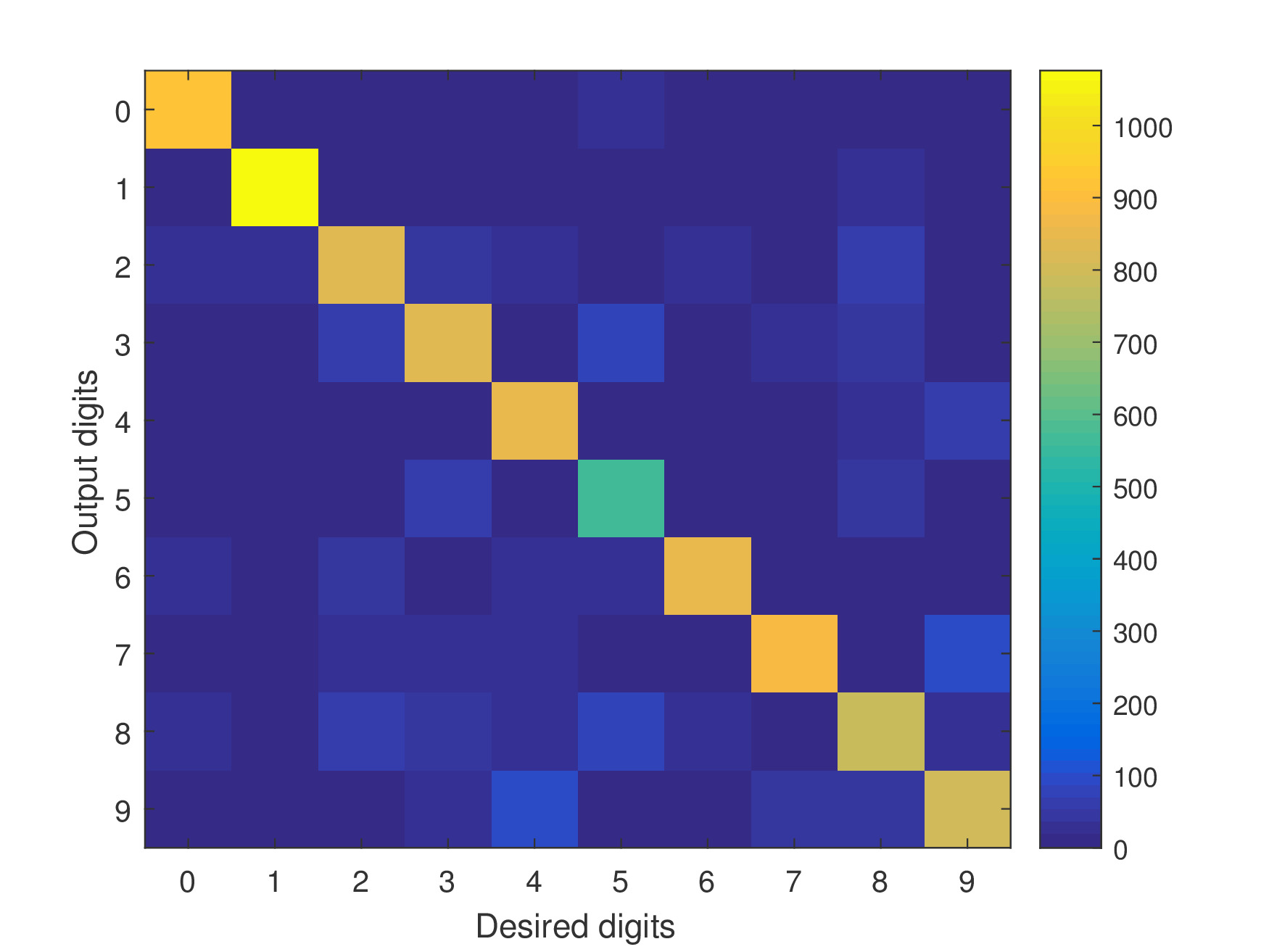}}
\hspace{\fill}
   \subfloat[\label{lr} ]{%
      \includegraphics[width=0.33\textwidth]{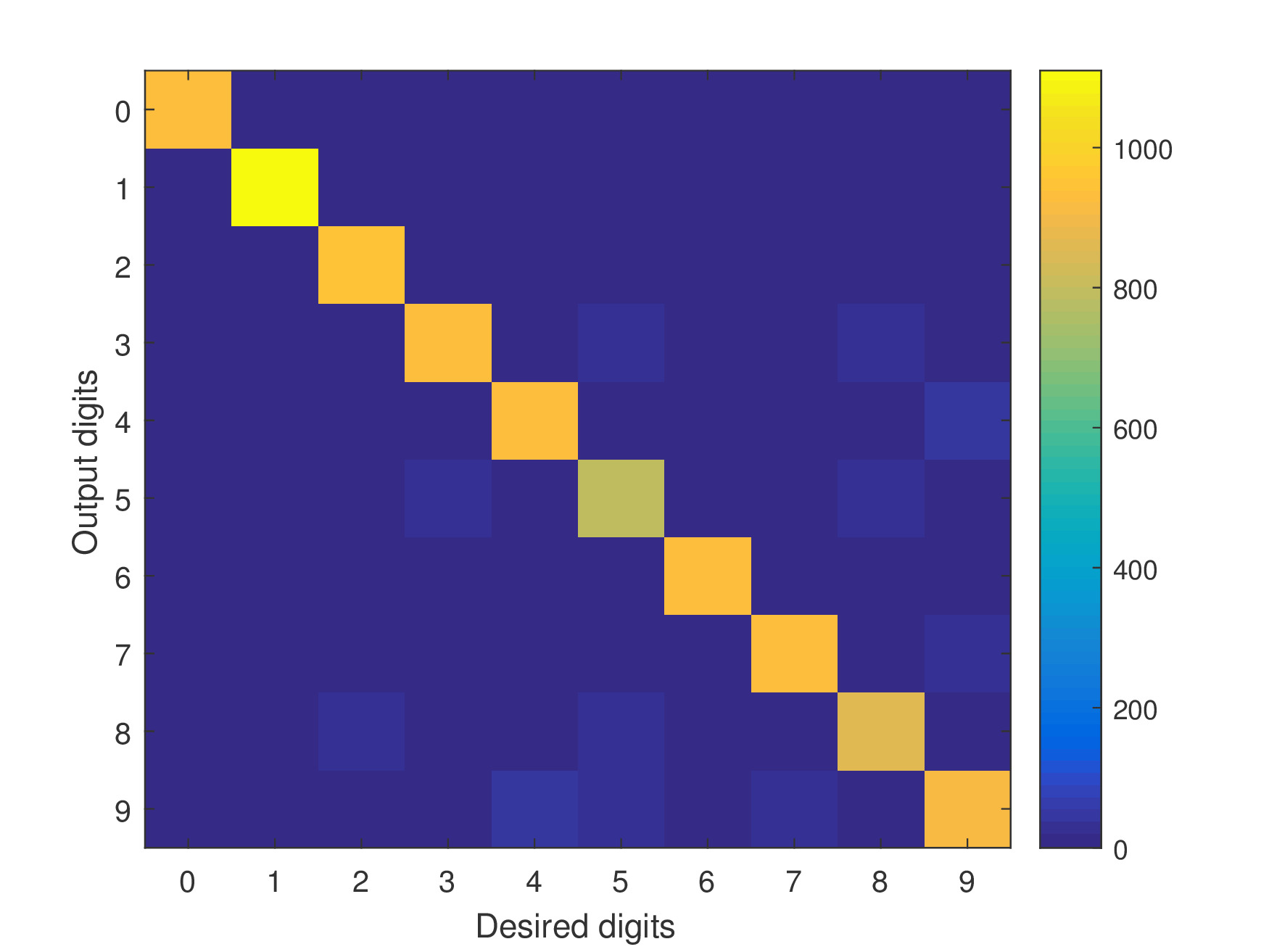}}
\hspace{\fill}
   \subfloat[\label{toolbox}]{%
      \includegraphics[width=0.33\textwidth]{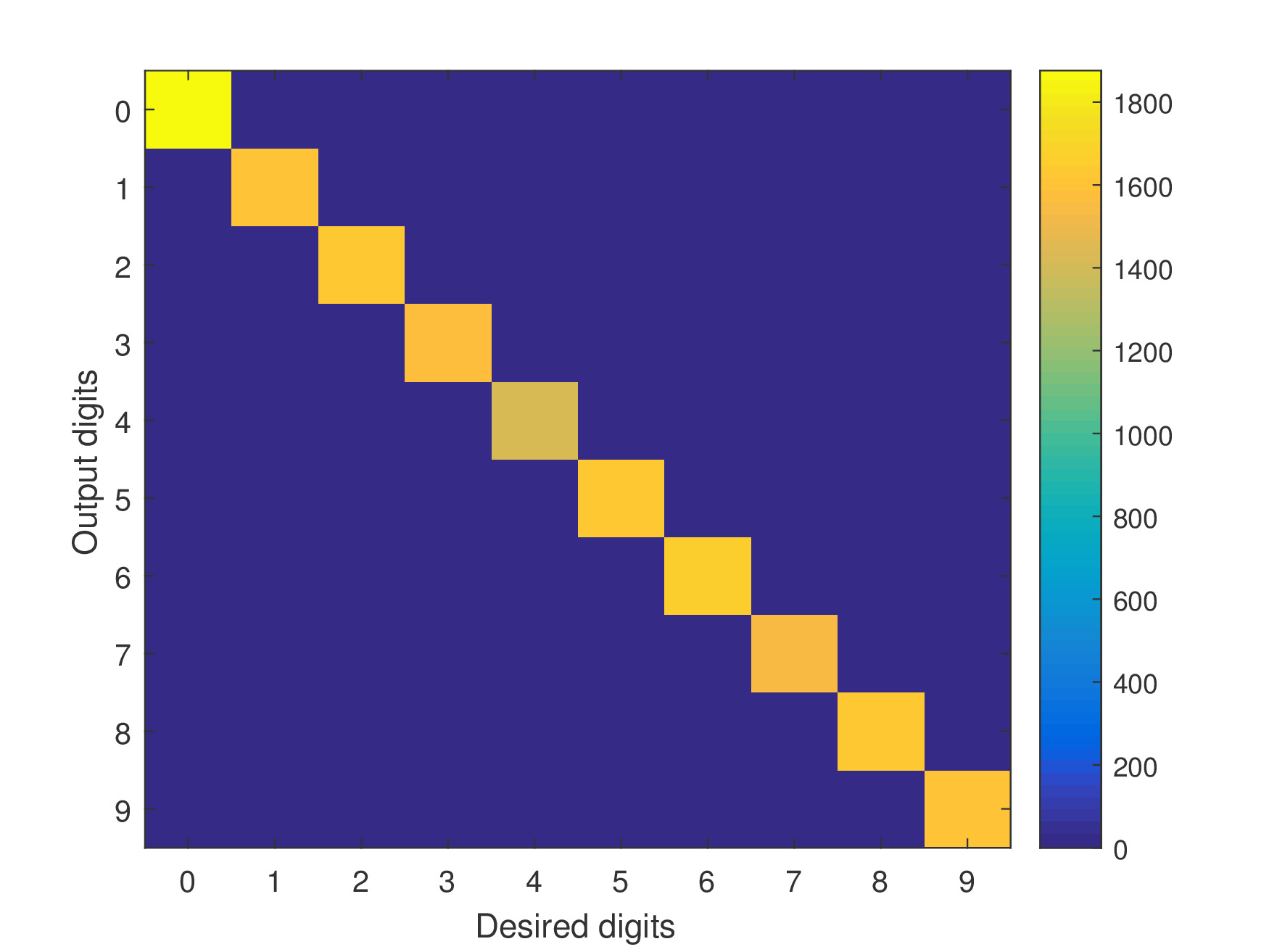}}\\
\caption{Average confusion matrix of the testing results presenting the obtained classification responses vs. the desired responses over ten presentations of the 10.000 standard MNIST test dataset. High values on the diagonal indicates correct estimation while at any other point indicates confusion between two digits. (a) STDP approach; (b) Simple Regression approach; (c) Scaled conjugate gradient back-propagation approach using Neural network training toolbox of Matlab 2018b.}\label{CM} 
\end{figure*}

\section{Results}
\textcolor{black}{\begin{figure}[!ht]
    \centering
    \includegraphics[width=0.45\textwidth]{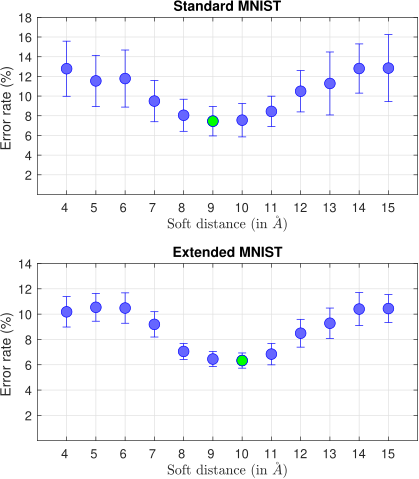}
    \caption{\textcolor{black}{The average (over ten benchmarks) network error rate (\%) and the average clustering coefficient of the network as a function of the soft distance, which assembles different connectivities. For each benchmark, different random initial conditions are used for each neuron along with different shuffle during the training phase.}}
    \label{performanceVSconnectivity}  
\end{figure}}

\textcolor{black}{We demonstrate the utility of the molecular-based RC architecture in unsupervised image classification by using the ReSuMe learning framework. We evaluated the proposed RC system with different RC connectivities for the standard MNIST and the extended MNIST datasets containing $60K/10K$ and $240K/40K$ of training/testing handwritten images, respectively. The connectivities, concerning the input layer to RC connectivity, the internal RC connectivity and the RC to Output layer connectivity, are initialized with random synaptic weights following the uniform distribution within the range $[0,1]$. Regarding the connections between the RC and the output layer, they are restricted during training within the range $[-1,1]$ to avoid bursting behaviors of the classifier neurons. To ensure the existence of the classifier's response (output spikes), we apply the Softmax activation function on the output neurons' membrane potentials ($S_o(t)\text{=softmax}(v_o(t))$). The ReSuMe learning method and the evaluation are applied right at the end of the learning duration and the training hyper-parameters are also optimised and set as follows: the learning duration $T\text{=}6\text{ms}$, the learning rate $\eta\text{=}10^{-5}$, the STDP amplitudes $A_{+}\text{=}A_{-}\text{=}1$ and the STDP time constants $\tau_{+}\text{=}\tau_{-}\text{=}5\text{ms}$. The dynamics of the Izhikevich model is evaluated by numerical integration with a time-step of $\Delta(t)\text{=}0.5$ms.}

\begin{table}[htpb]
    \centering
    \caption{\textcolor{black}{Overall Accuracy for different RC approaches.}}
    \begin{tabular}{ >{\color{black}}m{10em} >{\color{black}}c >{\color{black}}c} 
     \toprule \toprule
     \multirow{2}{*}{RC Approach} & \multicolumn{2}{c}{\textcolor{black}{Accuracy}}    \\ \cmidrule{2-3}
             & Standard MNIST & Extended MNIST\\ 
     \midrule
      No Connectivity   & 86.5\%   & 87.8\%    \\
      Hard Connectivity & 86.9\%   & 88.2\%    \\
      Soft Connectivity & 92.5\%   & 93.7\%    \\
     \bottomrule \bottomrule 
    \end{tabular} 
    \label{table:MyPerformances} 
\end{table}

\textcolor{black}{We evaluated three different RC dynamics caused by the proposed RC system without the internal connectivity, with the hard connectivity and the soft connectivity, explained in the previous section~\ref{ProposedModel}, to verify whether the molecular-based structure implies an overall accuracy enhancement. Except this, we performed for different soft distances, and for both MNIST-based datasets, 120 benchmark simulations, ten benchmarks for each soft distance considering the random initial parameters, to optimize the overall performance. The average RC error rate (\%) as a function of the soft distance is presented in Fig.~\ref{performanceVSconnectivity}. The accuracy for both datasets of the evaluated baseline RC system without any internal connectivity along with the accuracy of the proposed hard and the best soft internal connectivities are reported in Tab.~\ref{table:MyPerformances}. It is evident that by adopting a soft connectivity the dynamics of the LSM are affected, which eventually results in accuracy enhancement. Nevertheless, it should also noticed that after some point, a further increase of the soft distance leads to a loss of orthogonal connectivity, resulting in a decrease in the average accuracy achieved. The best accuracies we have managed to achieve for the MNIST classification task through the STDP-based ReSuMe learning framework is $92.5\%$ for the standard MNIST dataset (soft distance of $9\AA$), and $93.7\%$ for the extended MNIST dataset (soft distance of $10\AA$), which are clearly superior to that of the baseline accuracies ($86.5\%$ and $87.8\%$, accordingly). Fig.~\ref{performanceVSconnectivity} shows that there is a soft distance range in which the fractal-like orthogonal connectivity of the molecular structure emerge, as shown in Fig.~\ref{adjMATRIX}, and lower average error rates are obtained.} 

\begin{table*}[!htb]
    \centering
    \caption{Classification accuracy of different related SNNs and LSM-based architectures on MNIST-based test sets.}
    \begin{tabularx}{\linewidth}{@{}>{\color{black}}c*{10}{>{\color{black}}c}@{}}
    \toprule \toprule
     \multirow{3}{*}{Architecture} & \multirow{3}{*}{Network neurons} & \multirow{3}{*}{Network synapses} & \multicolumn{2}{c}{\textcolor{black}{Accuracy (\%)}} \\ \cmidrule{4-5}
     & & &  Standard MNIST & Extended MNIST \\
     
     \midrule
     Two-layered SNN [46] & $2\times6,400$ & 45,977,600    & 95.00 & - \\
     Four-liquid SpiLinC [47] & $4\times3,200$  & 4,866,048    & 90.90 & - \\
     Multiple-Liquid & \multirow{2}{*}{$4\times1,000$}   &  \multirow{2}{*}{400,000}   & \multirow{2}{*}{95.20} & \multirow{2}{*}{89.00}\\
     Multiple-Readouts RC [48] & & & &\\
     Molecular-based Liquid & \multirow{2}{*}{$1\times2,922$}   & \multirow{2}{*}{346,608}  & \multirow{2}{*}{92.50} & \multirow{2}{*}{93.70}\\ 
     (this paper) & & & &\\ 
     \bottomrule \bottomrule 
    \end{tabularx} \label{table:OtherPerformances} 
\end{table*}

\textcolor{black}{A comparison of related SNNs and LSM-based architectures used for the MNIST classification is shown in Tab.~\ref{table:OtherPerformances}. 
A network architecture similar to ours is presented in \cite{wijesinghe2019analysis}, which consists of four properly-scaled random ensembles that are made up of 1000 neurons each and manages to reach an almost identical 
accuracy by using the SGD training method regarding the standard MNIST dataset. However, 
the proposed molecular architecture yields better error-rates (percentage) trends than in \cite{wijesinghe2019analysis} work. Beyond that,} our work also utilizes about $27\%$ less trainable weights/synapses in the readout layer. While a full-scale training optimization, regarding different training approaches or stimulation handling techniques, of the proposed molecular-based LSM is yet to be demonstrated in our future work, the baseline accuracy achieved with very limited trainable weights might indicate the promising potential of the proposed topology. 

Despite the previous LSM-based approaches, there is also the SNN-based work\textcolor{black}{~\cite{Diehl2015}} that addresses the MNIST classification problem through a full-scale network training. Unlike the proposed LSM work, where training is performed on the readout layer,\textcolor{black}{\cite{Diehl2015}} perform training in the entire network of spiking neurons. Consequently, this work, by employing a single classifier, is considerably restricted to the number of trainable synaptic weights. Nevertheless, it should also considered that when similar SNN systems target for a hardware based implementation, such approaches request additional resources to control the overall process and at the same time require additional memory so as the newly calculated network weights can be stored. 

In Fig.~\ref{CM} average confusion matrices are shown under the Resume learning, the multiple linear regression and the SCG training accordingly. The results obtained present that the RC dynamics can be utilized through the RC state vector. This RC state consists of the overall spiking activity of each neuron. In this manner, by applying the multiple linear regression and the SCG training through the MATLAB's toolbox, a testing accuracy of $96.83\%$ and $98.66\%$ are achieved accordingly.

\section{Conclusions and Discussion}

We evaluated a computing potential of a neuromorphic network with architecture based on Verotoxin molecule by using a RC computing approach that hybridises computing on a single molecule and classification using neuromorphic models. We demonstrated an unconventional approach by assuming that each atom of a molecule is a neuron which produces action potentials by the Izhikevich neuron model and affects both its neigbouring neurons and the output layer. The output layer is trained by various training methods starting with the STDP-based ReSuMe training method, carrying on with the multiple linear regression and at last with the scaled conjugate gradient (SCG) back-propagation \textcolor{black}{to examine whether such a molecular-based protein topology can contribute in addressing neuromorphic challenges under the reservoir computing approach.} Despite the fact that we performed the training on a single readout layer, we managed to evaluate our proposed network \textcolor{black}{on the recognition of handwritten digits by the MNIST-based datasets, namely the standard MNIST dataset and the extended MNIST dataset,} and got a sufficient degree of accuracy \textcolor{black}{demonstrating more efficient performances than other similar LSM approaches with a single readout layer.} 

Our motivation initially emerges from the mechanisms that underlie the oscillatory behaviour of atoms in molecular structure to interact with their adjacent atoms. This phenomenon has inspired us and lead us to interpret atoms as biological neurons, which operate in the frequency domain. Thus, a molecular-based nanoelectronic device that would work on an atomic-scale through oscillations and could perform in neuromorphic applications~\cite{Bai2018, Pilati2019}, such as reservoir computing, has been envisaged. Although, for the time being, there is no such straightforward solution for easy implementation of such atomic-scale nanoelectronic devices owing to possible fabrication issues and corresponding difficulties, mainly attributed to the top-down approach of lithography process, and a fully secure bottom-up approach of creating complex molecular structures required to have even more steady molecular composition and limited defects~\cite{pandey2016aspects}, we propose here, as a first step, a promising protein structured reservoir computing approach that could eventually implement reservoir computing in the oscillation of actual protein molecules. \textcolor{black}{This oscillation can be practically realised for each molecular-coupled atom considered as an electron-spring system which can be modeled by harmonic oscillator models, like the classical Lorentz oscillator model. With the right conditions (electric fields, optical stimulation), we can directly affect atoms and cause excitations.} An excitation in a molecule takes place when an electron in a ground state absorbs a photon and moves up to high yet unstable energy level, which later returns to its ground state. When returning to the ground state, the electron releases photon which travels with speed $3\times 10^{18}$ \AA \, per second. Assuming at each step of modelling excitation wave-front travels $\rho=3$\AA, one step of the model evolution corresponds to one attosecond, of real time. \textcolor{black}{Considering atoms as neurons with two basic states, -- `excited' and `de-excited', a transition from `excited' to `de-excited' state corresponds to emission of spikes analogous to emission of photons as already discussed in literature~\cite{ezhov2000quantum,killoran2019continuous,gupta2001quantum}.} 
The outputs of the protein computing unit can be measured using controlled light waves and pulse trains~\cite{goulielmakis2008single,baltuvska2003attosecond,nabekawa2017probing, ciappina2017attosecond}, or in case of using less exotic devices, train of $10^3$ impulses of the same data inputs can be sent to the molecule and output recorded via accumulating, e.g. capacitive, devices. 
Therefore, we first intend to examine whether such a molecular-based protein topology for reservoir computing can contribute in addressing neuromorphic challenges, also demonstrating efficient performance, such as the recognition of handwritten digits by the MNIST dataset. Furthermore, we have presented that reservoir computing can take place with Izhikevich spiking neurons on the proposed molecular network topology. 

The MNIST classification problem on RC frameworks, every choice made concerning the control of Reservoir dynamics, the use of different kind of interconnected ensembles as well as the training approach of the readout layer has significant impact in the \textcolor{black}{overall} accuracy of the system. 
\textcolor{black}{We have managed to adapt this particular molecular structure, as performed in every other work related to reservoir computing, so as to achieve comparable or even better accuracies to other similar works.} A molecular-based architecture is designed to achieve comparable performances to other random and properly scaled connectivities. Nonetheless, its accuracy can be further enhanced by applying optimization techniques used by similar LSM implementations~\cite{wang2020software,wang2015general,wang2016d,zhou2019evolutionary,guo2020exploration,tian2020neural}. RC optimization techniques regarding the proposed topology can be illustrated with the control of the training procedure by using different learning methods, the investigation of different types of neurons to control the RC dynamics and the study of stimulation by utilizing various input interfaces or different stimulation times as frequently performed for such tasks.

The small-world effect with the high-clustering attribute of molecular connections classifies them as promising neuromorphic topologies for bio-inspired networks to implement high-dimensional transformations in neuromorphic computing architectures such as RC frameworks. Regarding the bio-inspired molecular ability to alter under different conditions they're imposed, could lead us to exploit their modification mechanisms or even to utilise the prospect of folding which could occur in real-time and investigate further the utilisation of such single-molecule systems.

\section*{Acknowledgment}

K.-A. T. gratefully acknowledges the financial support from the Hellenic Foundation for Research and Innovation (HFRI), under the HFRI PhD Fellowship grant (Fellowship No. 1228).

\bibliographystyle{unsrt}  
\bibliography{references}







\end{document}